%% file: main.tex
\documentclass{article}
\usepackage{iclr2027_conference,times}
\input{math_commands.tex}

\usepackage[colorlinks=true,allcolors=blue!60!black]{hyperref} %
\usepackage{url,booktabs,tabularx,array,amssymb,graphicx,xcolor,multirow}
\usepackage{longtable}

\usepackage[capitalize,nameinlink]{cleveref}
\crefname{section}{Sec.}{Secs.}
\crefname{appendix}{App.}{Apps.}
\crefname{figure}{Fig.}{Figs.}

\usepackage{enumitem}
\usepackage{wrapfig}
\usepackage{varwidth}
\newcommand{\promptbox}[1]{{\setlength{\fboxsep}{2pt}\fcolorbox{black!25}{black!4}{\begin{varwidth}[t]{0.78\textwidth}\itshape #1\end{varwidth}}}}

\makeatletter
\renewcommand\paragraph{\@startsection{paragraph}{4}{\z@}%
  {.25ex \@plus .1ex \@minus .1ex}%
  {-1em}%
  {\normalfont\normalsize\bfseries}}
\makeatother

\newcommand{\look}{\textsc{Look}}
\makeatletter\newcommand{\astable}{\def\@captype{table}}\makeatother
\definecolor{lookteal}{HTML}{147D83} %
\newcommand{\lookbox}[1]{\raisebox{0pt}[\ht\strutbox][\dp\strutbox]{\setlength{\fboxsep}{1.5pt}\setlength{\fboxrule}{0.8pt}\fcolorbox{lookteal}{white}{#1}}}
\usepackage{tikz}
\definecolor{looktint}{HTML}{E1EFEF} %
\DeclareRobustCommand{\chip}[3][none]{\raisebox{0pt}[\ht\strutbox][\dp\strutbox]{\tikz[baseline=(chip.base)]\node(chip)[fill=#2,draw=#1,line width=0.6pt,rounded corners=1pt,inner sep=1.5pt]{\strut #3};}}
\DeclareRobustCommand{\pnum}[1]{\raisebox{0pt}[\ht\strutbox][\dp\strutbox]{\tikz[baseline=(pnum.base)]\node(pnum)[circle,fill=lookteal,text=white,inner sep=0pt,minimum size=1.05em,font=\scriptsize\sffamily\bfseries]{#1};}}

\title{Learning to Look:\\A Foundation for Visual Understanding}
\author{
Wenshuai Zhao\textsuperscript{{\normalfont 1,2}}\enspace
Yihao Wang\textsuperscript{{\normalfont 1,2}}\enspace
Benjamin Robson\textsuperscript{{\normalfont 1,2}}\enspace
Juho Kannala\textsuperscript{{\normalfont 1,3}}\enspace
Arno Solin\textsuperscript{{\normalfont 1,2}}\\
\normalfont\textsuperscript{1}Aalto University\enspace
\textsuperscript{2}ELLIS Institute Finland\enspace
\textsuperscript{3}University of Oulu}
\iclrpreprintcopy
\hypersetup{
  pdftitle={Learning to Look: A Foundation for Visual Understanding},
  pdfauthor={Wenshuai Zhao, Yihao Wang, Benjamin Robson, Juho Kannala, Arno Solin}
}
\begin{document}
\maketitle

\begin{abstract}
Vision-language models (VLMs) suffer from visual hallucination, such as saying an object that is not present in an image. We ask a more fundamental question: whether a VLM is even aware of where it is looking before understanding the semantic meaning in the image. We formulate this question as a \look{} task: identifying which parts of its visual input were available to the model. We construct the task in a self-supervised way without manually labeling the available region to the model. Our experiments show that existing VLMs fail to localize the visible region. We further show that fine-tuning VLMs on the proposed \look{} task improves a model's awareness of visual input.

\end{abstract}
\begin{center}
\small Project page: \url{https://aaltoml.github.io/look/}
\end{center}

\begin{figure}[!b]
\centering
\includegraphics[width=\textwidth]{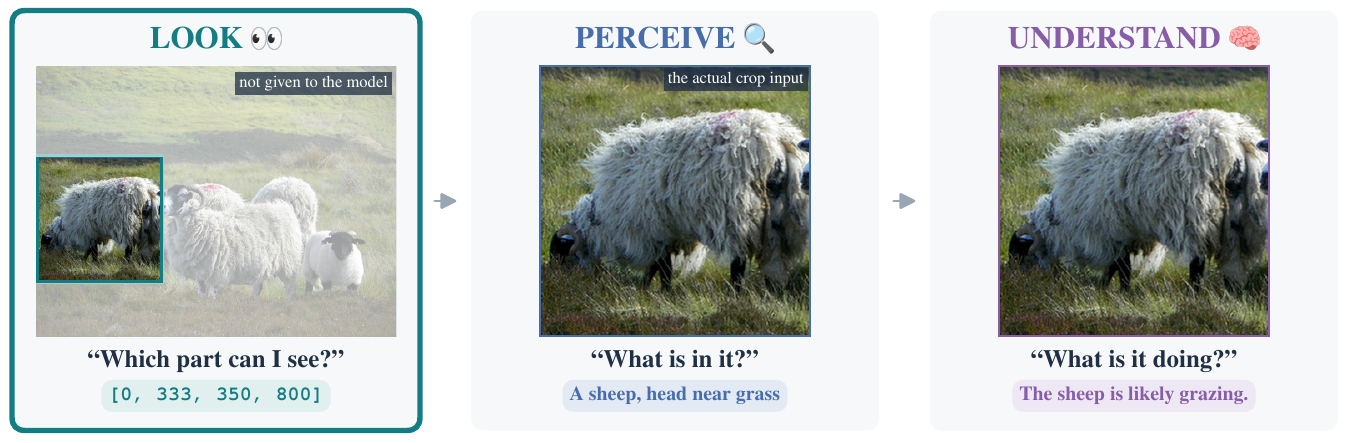}\par
\vspace*{-6pt}
\caption{\textbf{Hierarchy of visual understanding: \lookbox{\look{}} $\rightarrow$ Perceive $\rightarrow$ Understand.} 
We introduce a hierarchical visual grounding structure and argue that visual hallucination can arise from any level. This work focuses on the fundamental \textit{Look} level.
}
\label{fig:teaser}
\end{figure}

\section{Introduction}
\label{sec:intro}

Vision-language models (VLMs) can describe images fluently while referring to objects that are not actually present. This failure mode, commonly termed visual hallucination, has been widely evidenced~\citep{rohrbach2018chair,li2023pope,chen2024rope} and persists even in frontier models. For example, \citet{asadi2026mirage} show that Gemini-3-Pro, Gemini-2.5-Pro, Claude-Opus-4.5, and GPT-5.1 can readily produce detailed image descriptions and elaborate reasoning traces even when no image is provided. This observation suggests that visual hallucination can arise prior to visual misinterpretation: the model may fail to reliably distinguish between the presence and absence of visual input. In this sense, current VLMs lack a basic form of visual-input awareness, the ability to recognize whether they are receiving visual information at all. This stands in sharp contrast to human visual perception, where an observer is inherently aware of whether visual input is available, for example, whether their eyes are open or closed.

However, existing efforts to address visual hallucination are limited at the level of semantics. They curate instruction data with negative and counterfactual examples~\citep{liu2024lrv}, align outputs with factual or human preferences~\citep{sun2024llavarlhf}, or intervene during decoding to suppress content the image does not support~\citep{leng2024vcd,huang2024opera}. Despite their differences, these methods address essentially the same question: whether the model's claims about the visual scene are semantically correct. We instead ask a more fundamental question that precedes semantic interpretation: \textcolor{orange}{\emph{Is a VLM aware of where it is looking in the first place before reasoning about what the visual input means?}}

To this end, we argue that general visual grounding can be organized into a three-level hierarchy as shown in~\cref{fig:teaser}: \emph{Look} identifies where the model is looking, \emph{Perceive} determines what is present there, and \emph{Understand} uses this evidence to answer a question or draw an inference. We hypothesize that visual hallucination may arise as early as the first level, \emph{Look}, a capability that remains largely untested by existing benchmarks. A model should ground its claims in available visual input when asserting that it sees something and, more fundamentally, should be aware of whether visual input is available at all. 

We therefore propose \look{}, a task designed to probe a VLM's awareness of the visual input available to it. Specifically, the model is given an image crop and must localize the visible region as a bounding box in the coordinate frame of the original image. We encode only the crop pixels, while assigning each crop patch the same positional embedding it would have received in the full image. Thus, although the excluded scene content is unavailable, the model retains all geometric information required to localize the crop. Intuitively, this is analogous to covering one of a person's eyes and asking them to identify which eye is covered. \look{} probes a model's awareness of the visual input available to it, rather than the semantic content of what it sees.

Our experiments primarily use Qwen3-VL, which preserves absolute positional information for image patches, whereas many other VLMs rely only on relative positional embeddings and therefore lack the spatial cues required for the \look{} task. We first show that pretrained VLMs fail to reliably localize the visual input available to them. We then fine-tune Qwen3-VL on automatically constructed \look{} data and find that the model can learn the task and generalize well to unseen examples. To evaluate whether \look{} improves a VLM's awareness of visual input more broadly, we further introduce the Visual Awareness Benchmark (VisABench). In VisABench, visual tokens are replaced with null \texttt{\textless image\_pad\textgreater} tokens, and the model is asked whether a target object is present in the image. Models fine-tuned on \look{} substantially outperform the corresponding baselines, suggesting that learning to \look{} enhances a model's awareness of available visual evidence and naturally helps mitigate visual hallucination.

Our contributions are threefold:
\begin{enumerate}[
    label=(\roman*),
    labelindent=0pt,
    leftmargin=*,
    labelsep=0.5em,
    align=left,
    widest=iii
]
    \item We introduce a three-level hierarchy of visual grounding. The hierarchy decomposes visual grounding into \emph{Look}, \emph{Perceive}, and \emph{Understand}, providing a framework for studying visual hallucination at a more fundamental level.

    \item We propose the \look{} task for improving visual awareness through self-supervision. \look{} training data can be generated from arbitrary image datasets by cropping images, with the crop region itself providing bounding-box supervision and requiring no semantic annotation.

    \item We introduce VisABench to evaluate a VLM's awareness of visual input. We show that pretrained VLMs can hallucinate object presence even when visual evidence is absent, while fine-tuning on \look{} substantially improves visual awareness and reduces such hallucinations.
\end{enumerate}

\section{Hierarchy of Visual Understanding}
\label{sec:hierarchy}
Visual understanding is commonly evaluated by the success rate on visual question answering (VQA) tasks. However, when a model produces an incorrect answer, it is often difficult to determine where the failure originates. We argue that visual understanding involves multiple stages, and that visual hallucination can arise at any of them. Effectively mitigating visual hallucination therefore requires identifying the stage at which the failure occurs and addressing its underlying cause.

To this end, we propose a three-level hierarchy of visual understanding as shown in~\cref{fig:teaser}. Let $I$ denote an image, $w$ a region of it, $I_w$ the pixels inside $w$, and $P_w$ the positions those pixels occupy in $I$. The three levels answer three different questions about the given image input:
\begin{equation}
\underbrace{\hat w = L(I_w,P_w)}_{\text{\emph{Look}: did I see and where is it?}},\qquad
\underbrace{\hat z = P(I_w)}_{\text{\emph{Perceive}: what is in it?}},\qquad
\underbrace{\hat a = U(\hat z,q)}_{\text{\emph{Understand}: what follows?}},
\label{eq:hierarchy}
\end{equation}
where $\hat z$ denotes observations and $q$ a question. Particularly, the ordering of the hierarchy expresses the dependence of evidence to conclude the final visual understanding: a claim about what the image contains ($P$) or implies ($U$) should presuppose some account of what input was available ($L$). Consider the limiting case where an input that carries no visual content at all. In this case, the correct output of $L$ is that nothing is visible, and every positive claim of $P$ is a hallucination without grounding~\citep{asadi2026mirage}. 

\begin{figure}[t]
\centering
\includegraphics[width=\textwidth]{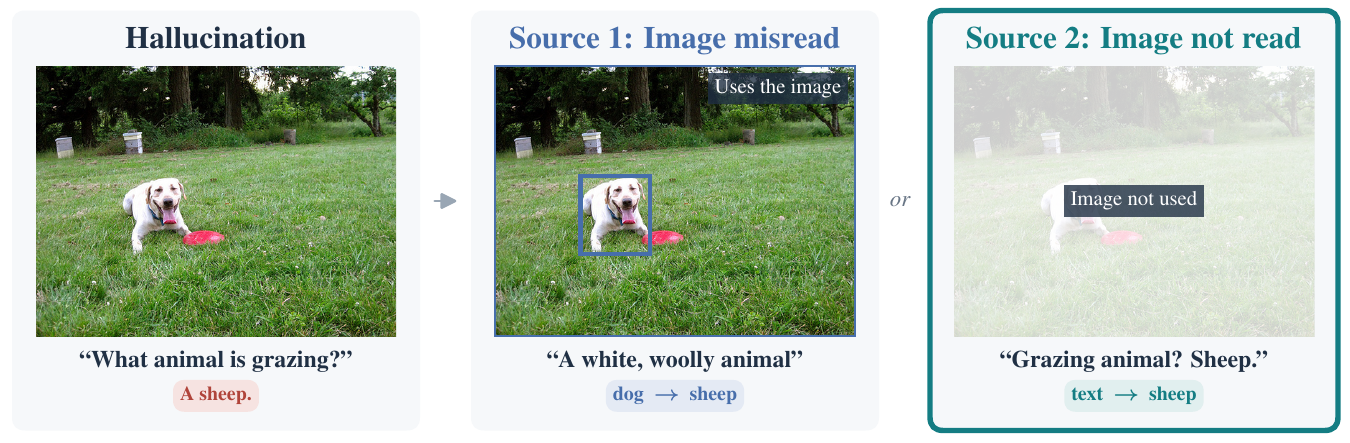}\par
\vspace*{-6pt}
\caption{\textbf{Where does a hallucination originate?} In this example, we ask what animal is grazing in a photograph of a dog on grassland, a model answers ``a sheep''. The same wrong answer can arise at two levels of the hierarchy. At the \emph{Perceive} level, the model uses the image but misrecognizes the dog. At the \emph{Look} level, the model does not rely on the image at all, and ``sheep'' follows from the textual context.}
\label{fig:failure}
\end{figure}

\paragraph{Source of visual hallucination.} While existing work largely focuses on semantic-level visual perception and reasoning, our hierarchy highlights the more fundamental role of \emph{Look} in visual grounding. Before interpreting visual content, a model should first be aware of the visual input available to it and distinguish visual evidence from textual context. For example (\cref{fig:failure}), if a model hallucinates a \textit{sheep} in an image of grassland containing a \textit{dog}, the failure may arise because it misperceives the dog as a sheep, corresponding to a failure at the \emph{Perceive} level. More fundamentally, however, the model may fail to rely on the image at all and instead generate \textit{sheep} from textual associations with the surrounding context, corresponding to a failure at the \emph{Look} level. Distinguishing these failure modes is essential for identifying where visual hallucination originates and how it should be addressed.\looseness-1

\section{The \look{} task}
\label{sec:task}

Motivated by the proposed hierarchy and the goal of improving visual grounding at the fundamental \emph{Look} level, we design the \look{} task to quantify a VLM's awareness of its available visual input. We further use the task as a self-supervised training objective to enhance this awareness. A central challenge, however, is to disentangle visual-input awareness from semantic shortcuts that could allow the model to solve the task without actually identifying where it is looking.

\subsection{Constructing a \look{} example}
\label{sec:positions}
\begin{figure}[t]
\centering
\includegraphics[width=\textwidth]{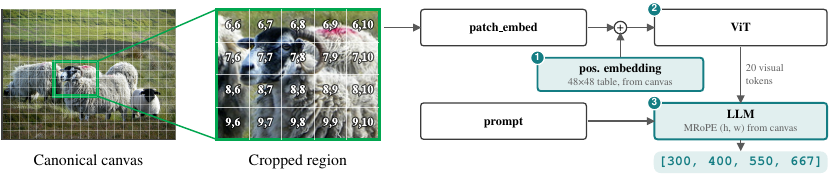}\par
\vspace*{-6pt}
\caption{\textbf{Constructing a \look{} example.} We select a merge-block-aligned viewport and encode only its pixels for the VLM. Importantly, we restore positional metadata from the full canvas without encoding the excluded image content. The crop's coordinates provide the target automatically. \chip[lookteal]{looktint}{Changed} blocks indicate places where token positions are added.}
\label{fig:method}
\end{figure}

Just as a person knows which part of their surroundings falls within their view, a model should know which part of an image it receives. In the \look{} task, the full image plays the role of the environment and a crop plays the role of the available view (\cref{fig:method}). We sample a rectangular window aligned with the model's visual-token grid, encode only the pixels inside it, and ask the model for the bounding box of the window in the coordinate frame of the full image. Because we sample the window ourselves, the label can be automatically obtained and any image can be used as \look{} examples.

However, a crop encoded on its own carries no trace of where it came from. The same pixels could occur anywhere in a larger image, and a model could only guess from semantic cues, such as sky suggesting the top of a scene. We therefore assign each crop token the positional information it would have received in the full image. We implement \look{} on Qwen3-VL~\citep{qwen3vl2025}, which preserves absolute positional information for image patches and thus provides the spatial cues needed to identify the available region. Specifically, as illustrated in~\cref{fig:method}, we preserve the crop's original full-image positions at two points in the architecture: the vision encoder's absolute positional embeddings~\pnum{1} and the language model's multimodal rotary positional embeddings (MRoPE)~\pnum{3}. In this way, the geometric information required for localization is explicitly available, allowing the model to identify its available visual region without relying on semantic cues to infer its location. The \look{} task therefore directly probes whether the model can recognize where its visual input comes from. Implementation details are provided in \cref{app:implementation}.

\subsection{Training with \look{} data}
\label{sec:training}
\paragraph{Visual perception verbs as the interface between language and vision.} To improve a model's visual awareness, we encourage it to be sensitive when generating expressions that explicitly invoke visual perception, such as \texttt{I see} and \texttt{I observe}. We view such expressions as a linguistic interface between the language and visual modalities: when a model claims to see something, that claim should be grounded in the visual input available to it. Accordingly, we equip each \look{} example with diverse localization prompts that invoke such visual-perception expressions, as listed in~\cref{app:prompts}.

Concretely, each \look{} example pairs a localization prompt $q_{\rm look}$ with the bounding box $y(w)$ of the sampled window, represented in the model's coordinate format as defined in~\cref{eq:target}. Given an existing instruction-tuning mixture dataset $D_{\rm inst}$ and the generated \look{} dataset $D_{\rm look}$, we train on their union using the standard autoregressive cross-entropy objective:
\begin{equation}
\mathcal L(\theta)=-\frac{1}{\sum_{(x,q,r)\in D}|r|}\sum_{(x,q,r)\in D}\,\sum_{t=1}^{|r|}\log p_\theta(r_t\mid x,q,r_{<t}),
\qquad D=D_{\rm inst}\cup D_{\rm look},
\label{eq:loss}
\end{equation}
where $r$ is the target response of an example and $|r|$ its length in tokens, so the loss is averaged over all response tokens. For \look{} examples, $r$ is the serialized box $y(w)$ and $x=(I_w,P_w)$ denotes the cropped visual input and its positional information, while ordinary instruction examples retain their original image inputs and responses.

\section{VisABench: Evaluating VLMs' Illusion}
\label{sec:visabench}
The \look{} task examines where the available visual input lies within a scene. Its limiting case concerns whether any visual input is available at all. A model that is aware of its visual input should recognize when none is present, rather than describe content it has never received, the \textit{illusion} failure shown in \citet{asadi2026mirage}. To evaluate VLMs' illusion, we introduce the Visual Awareness Benchmark (VisABench).

\begin{figure}[t]
\centering
\includegraphics[width=0.8407\textwidth]{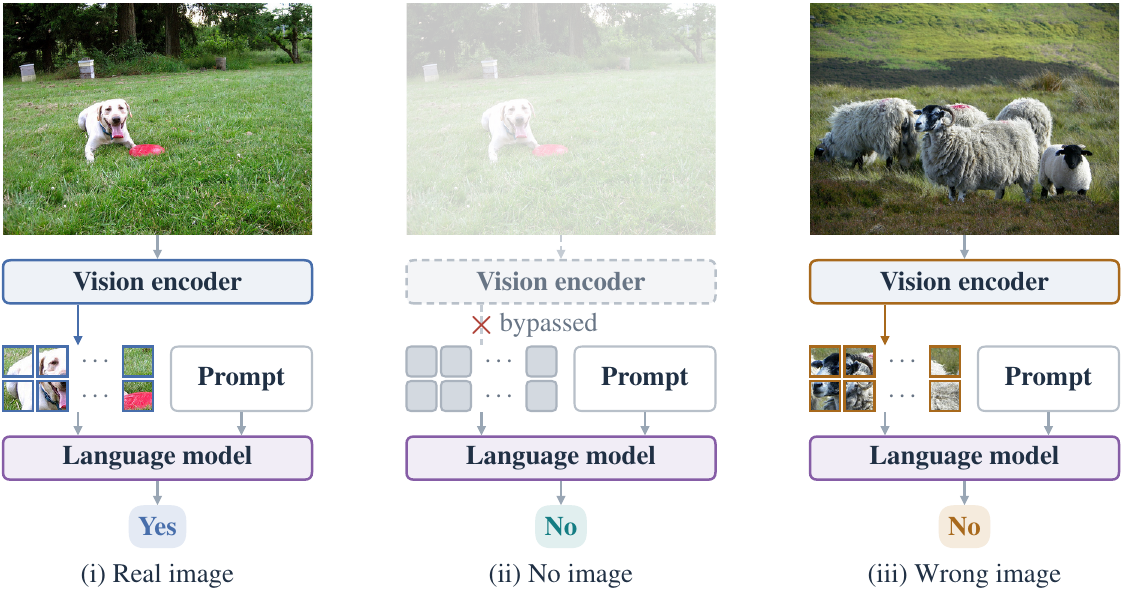}\par
\vspace*{-6pt}
\caption{\textbf{Illustration of VisABench construction.} All conditions use the same prompt, ``\textit{Does this image contain: dog? Answer with Yes or No.}'', with the only difference being the visual input. The grounded answer is Yes in (i) and No in (ii) and (iii).}
\label{fig:visa_method}
\end{figure}

\paragraph{Benchmark construction.}
VisABench uses all $10{,}752$ referring expressions from the RefCOCO testA and testB splits~\citep{yu2016refcoco}, spanning $1{,}500$ COCO images~\citep{lin2014coco}. For each expression, we ask the model: ``\textit{Does this image contain: \{expression\}? Answer with Yes or No.}'' We evaluate three input conditions, illustrated in~\cref{fig:visa_method}:
\begin{enumerate}[label=(\roman*),leftmargin=*,nosep]
\item \emph{Real image.} The expression is paired with its original image, which contains the referred object; the correct answer is therefore Yes.
\item \emph{No image.} We bypass the vision encoder while retaining the \texttt{<|image\_pad|>} placeholder embedding at each visual-token position. This preserves the sequence length, token positions, and attention mask of the corresponding real-image input while removing all visual content. The correct answer is therefore No.
\item \emph{Wrong image.} Each image is paired with the expression of a different item, for which the expected answer is No.
\end{enumerate}
We use the same $10{,}752$ items, the same set of images and referring expressions throughout for the three conditions. The no-image condition serves as our primary test of visual awareness.

\paragraph{Metrics.}
We score the model from the logits $\ell$ at the first answer position instead of parsing generated text, since free-form responses can vary substantially. Answers such as ``Yes.'', ``yes, it does'', or a longer sentence that never clearly commits to either answer make string-based evaluation brittle and discard information about confidence. We therefore compute
\begin{equation}
p(\text{Yes})=\sigma\Big(\log\sum_{t\in\mathcal Y}e^{\ell_t}-\log\sum_{t\in\mathcal N}e^{\ell_t}\Big),
\label{eq:pyes}
\end{equation}
where $\sigma$ is the logistic function and $\mathcal Y$ and $\mathcal N$ contain every vocabulary token whose surface form, ignoring case and surrounding whitespace, is Yes and No, respectively. We classify an answer as Yes when $p(\text{Yes})>0.5$. This readout requires only a single forward pass, avoids parsing failures, and provides a continuous confidence score.

We report the following metrics for each of the three input conditions:
\begin{enumerate}[label=(\arabic*),leftmargin=*,nosep]
\item \emph{Yes rate}: the fraction of examples for which the model answers Yes. In the no-image condition it is our primary metric ($\downarrow$), corresponding to false claims that the referred object is visually present. In the real-image condition ($\uparrow$) it controls for whether the model still recognizes the referred object when the correct image is provided; a model that always answers No would score perfectly on the no-image Yes rate but fail here. In the wrong-image condition ($\downarrow$) it gives a coarse measure of whether the model checks the referring expression against the actual image content.
\item \emph{Mean $p(\text{Yes})$}: the average probability assigned to Yes, with the same direction as the Yes rate in each condition. It captures the model's confidence even when this remains below the decision threshold.
\item \emph{Gap} ($\uparrow$): the difference between the real-image and no-image Yes rates, summarizing the model's sensitivity to the presence of visual evidence.
\end{enumerate}

\section{Experiments}
\label{sec:experiments}
\begin{table}[t]
\centering\small\setlength{\tabcolsep}{6pt}
\caption{\textbf{Compare different probe methods for visual awareness.} We report IoU of the pretrained Qwen3-VL when asked for the visible region over $1{,}000$ views. IoU counts answers without a valid box as zero. Methods operating on pixels or patch embeddings fail to probe visual awareness because the model can exploit semantic shortcuts, often predicting the bounding box of the dominant object rather than the available visual region.}
\label{tab:probes}
\begin{tabular}{lccccccc}
\toprule
Probe methods & Full image & \multicolumn{2}{c}{Pixel operation} & \multicolumn{2}{c}{Patch embedding} & \multicolumn{2}{c}{Crop}\\
\cmidrule(lr){3-4}\cmidrule(lr){5-6}\cmidrule(lr){7-8}
 & & Black & Grey & Zero & Mean & Local pos. & Global pos. (\look{})\\
\midrule
IoU $\downarrow$ & 0.52 & 0.75 & 0.77 & 0.58 & 0.76 & \textbf{0.18} & \textbf{0.18}\\
\bottomrule
\end{tabular}
\end{table}

\begin{figure}[t]
\centering
\includegraphics[width=\textwidth]{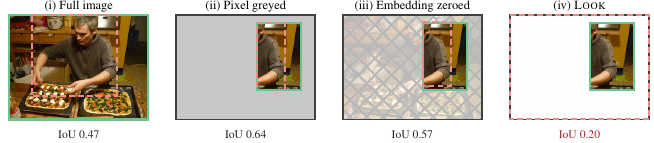}\par
\vspace*{-6pt}
\caption{\textbf{Four ways to probe visual awareness.} The green box shows the available visual region while the red dashed box shows the prediction. (i) We input the full image, but the model boxes the dominant person rather than the whole canvas. (ii, iii) When the context pixels are greyed out or its embeddings are zeroed, the model still locates the dominant person, likely using semantic shortcuts. (iv) In \look{}, the model answers with the whole canvas.}
\label{fig:ladder}
\end{figure}

We design experiments to answer four questions: (i) whether \look{} measures visual awareness without semantic shortcuts, (ii) whether the \look{} task is learnable, (iii) whether \look{} fine-tuning improves VisABench performance, and (iv) whether \look{} fine-tuning preserves the model's general visual perception capabilities.

\subsection{Setup}
\label{sec:setup}
\paragraph{Training dataset.}

To improve visual awareness while preserving the VLM's general capabilities, we train on a mixture of replay data and the proposed \look{} data. (i) Replay mixture dataset. We use $46{,}000$ instruction-tuning examples drawn from eight public datasets as detailed in \cref{app:training}. (ii) \look{} dataset. We construct the \look{} data from the same replay images by sampling one window per image as described in \cref{sec:task} and pairing each crop with one of the $36$ localization prompts listed in \cref{app:prompts}. To study the effect of \look{} supervision, we combine the full replay mixture with $10\%$, $20\%$, and $100\%$ of the \look{} dataset, denoted as \look{}-to-replay ratios of $1{:}10$, $2{:}10$, and $10{:}10$.

\paragraph{Model training and baselines.}
We experiment with Qwen3-VL-8B-Instruct~\citep{qwen3vl2025} since it uses absolute positional embeddings for image patches. We fine-tune the model with LoRA~\citep{hu2022lora} of rank $16$ and scale $32$ on all linear layers, keeping the vision encoder and the projector frozen, and train for two epochs with a learning rate of $10^{-4}$, an effective batch size of $32$, and a maximum sequence length of $4096$, using the objective in \cref{eq:loss}. Training is implemented with \textit{ms-swift}~\citep{zhao2025swift} on four H200 GPUs. We compare three configurations. (i) The \emph{SFT} baseline is trained on the replay instruction mixture alone. (ii) \emph{Ours} is trained on the mixture together with \look{} data at the three ratios above. (iii) \emph{LocPos} is an ablation that uses the same \look{} data as Ours (10:10) but disables the positional restoration of \cref{sec:positions} during training, so that each crop carries crop-local positions. This ablation separates the contribution of global positional information from that of the crop data and the answer format.

\subsection{How to measure visual awareness without semantic shortcuts?}
\label{sec:probes}
We aim to measure a VLM's awareness of its available visual input while preventing it from exploiting semantic shortcuts. We compare four families of probes as illustrated in \cref{fig:ladder}. (i) Full image. We input the full image and ask the model about its visible region. (ii) Pixel operations. We paint the hidden region black or gray. (iii) Patch-embedding operations. We set the embeddings of the hidden patches to zero or to their mean. (iv) Crop based. We encode only the visible region, with either local or global positions used to compute the absolute positional embeddings in Qwen3-VL. 

We evaluate each probe on the pretrained model using $1{,}000$ views sampled from the RefCOCO validation images~\citep{yu2016refcoco,lin2014coco}. As shown in \cref{tab:probes}, pixel- and patch-embedding-based probes achieve higher IoU, but this performance is largely driven by semantic shortcuts. The model often predicts the extent of dominant objects rather than the available visual region, a behavior also evident in the full-image setting. We therefore adopt cropping with global positional information for the \look{} task, as it removes the hidden content while preserving the geometric information required for localization.

\begin{table}[t]
\centering
\caption{\textbf{Performance on \look{} task.} Results are tested on the same $1{,}000$ views of \cref{tab:probes}. Success@$0.5$ is the fraction of views with an IoU of at least $0.5$.}
\label{tab:look}
\small
\begin{tabular}{lcccccc}
\toprule
Method & Base & SFT & LocPos & \multicolumn{3}{c}{Ours}\\
\cmidrule(lr){5-7}
 & & & & 1:10 & 2:10 & 10:10\\
\midrule
IoU $\uparrow$ & 0.181 & 0.111 & 0.371 & 0.609 & 0.665 & \textbf{0.742} \\
Success@$0.5$ $\uparrow$ & 0.025 & 0.012 & 0.322 & 0.734 & 0.808 & \textbf{0.844} \\
\bottomrule
\end{tabular}
\end{table}

\subsection{Learning to LOOK}
\label{sec:lookresults}

\begin{wrapfigure}{r}{0.40\textwidth}
\vspace{-12pt}
\centering
\includegraphics[width=0.40\textwidth]{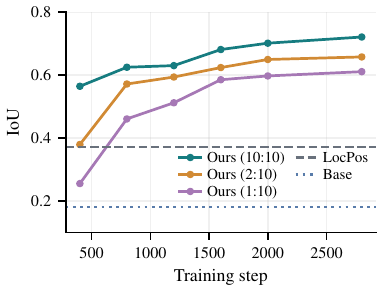}
\vspace{-18pt}
\caption{Learning to look dynamics.}
\label{fig:learning}
\vspace{-10pt}
\end{wrapfigure}
In this section,  we examine whether the \look{} task is learnable. Following the training setup in~\cref{sec:setup}, we compare model performance on the \look{} task. As the results shown in \cref{tab:look} and the learning dynamics in~\cref{fig:learning}, combining replay instruction data with \look{} data substantially improves localization performance. In contrast, LocPos fails to learn the task because it lacks the global positional information used by our method. Notably, even a $1{:}10$ ratio of \look{} data to replay data yields a substantial improvement.

\begin{figure}[t]
\centering
\includegraphics[width=\textwidth]{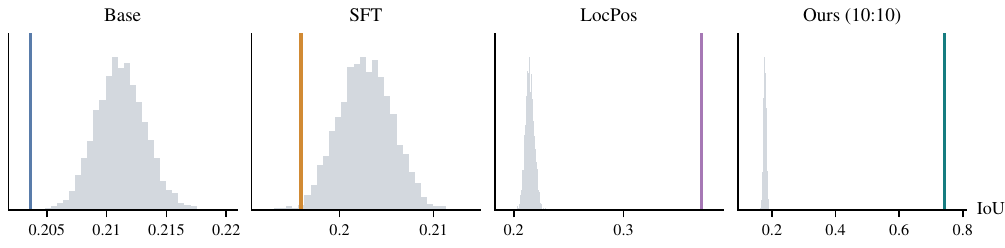}\par
\vspace*{-6pt}
\caption{\textbf{Permutation test.} The grey bars mean IoU after re-pairing each prediction with the window of another image while the color lines show the IoU on the given image. The pretrained and SFT models show no positive dependence on the view.}
\label{fig:permutation}
\end{figure}

\paragraph{Do predictions depend on the given view?} To further assess if the model is truly aware of its visual input, we perform a permutation test by pairing each prediction with the target window from another image. As shown in~\cref{fig:permutation}, the Base and SFT models achieve even higher IoU under permutation (grey bars), indicating that their predictions are poorly grounded in the actual visual input.

\subsection{Does \look{} fine-tuning improve VisABench?}
\label{sec:visaresults}
\begin{table}[t]
\centering
\caption{Performance on \textbf{VisABench.} All models are evaluated on the same $10{,}752$ items. For each condition, we report the Yes rate (\%) and mean $p(\text{Yes})$. \emph{Gap} denotes the real-image Yes rate minus the no-image Yes rate. The bottom rows report results for additional pretrained VLMs~\citep{bai2025qwen25vl,liu2024llava15,gemma2026gemma4}. $^\dagger$ The image-placeholder embedding of this model is untrained, causing it to produce neither Yes nor No without an image. We therefore omit its no-image results.}
\label{tab:visa}
\small\setlength{\tabcolsep}{6.8pt}
\begin{tabular}{lcccccccc}
\toprule
& \multicolumn{2}{c}{No image} & \multicolumn{2}{c}{Real image} & \multicolumn{2}{c}{Wrong image} & \\
\cmidrule(lr){2-3}\cmidrule(lr){4-5}\cmidrule(lr){6-7}
& Yes (\%) $\downarrow$ & $p(\text{Yes})$ $\downarrow$ & Yes (\%) $\uparrow$ & $p(\text{Yes})$ $\uparrow$ & Yes (\%) $\downarrow$ & $p(\text{Yes})$ $\downarrow$ & Gap $\uparrow$\\
\midrule
Base & 2.5 & \textbf{0.041} & 93.3 & \textbf{0.927} & 15.6 & 0.158 & 90.7 \\
SFT & 31.2 & 0.409 & 94.9 & 0.915 & 17.0 & 0.196 & 63.7 \\
LocPos & 5.7 & 0.272 & 92.6 & 0.890 & 15.0 & 0.170 & 86.9 \\
Ours (1:10) & 2.3 & 0.202 & 94.1 & 0.910 & 15.7 & 0.183 & 91.8 \\
Ours (2:10) & 4.9 & 0.241 & 93.9 & 0.911 & 16.0 & 0.188 & 89.0 \\
Ours (10:10) & \textbf{0.8} & 0.101 & 94.1 & 0.909 & 15.1 & 0.175 & \textbf{93.2} \\
\midrule
Qwen2.5-VL-7B$^\dagger$ & -- & -- & 90.0 & 0.868 & 12.5 & 0.129 & -- \\
LLaVA-1.5-7B$^\dagger$ & -- & -- & \textbf{95.3} & 0.775 & 30.2 & 0.355 & -- \\
Gemma-4-12B & 42.9 & 0.434 & 88.2 & 0.880 & \textbf{12.2} & \textbf{0.123} & 45.4 \\
\bottomrule
\end{tabular}
\end{table}

We next evaluate whether training on \look{} improves visual awareness as measured by VisABench. As shown in~\cref{tab:visa}, \look{} fine-tuning consistently improves performance. In particular, Ours (10:10) reduces the no-image Yes rate to $0.8\%$, compared with $2.5\%$ for the pretrained model and $31.2\%$ for standard SFT, while maintaining a comparable real-image Yes rate. It also achieves the largest real-to-no-image gap of $93.2$. These results indicate that \look{} improves the model's awareness of whether visual input is actually available without compromising its ability to recognize objects when an image is present. We note that this failure is also not specific to Qwen3-VL. For example, Gemma-4-12B~\citep{gemma2026gemma4} answers Yes on $42.9\%$ of no-image items.

\subsection{Does \look{} fine-tuning preserve general capabilities?}
\label{sec:general}
\begin{table}[t]
\centering\small
\caption{\textbf{\look{} maintains general capabilities.} We report accuracy (\%) on MMStar, MMMU and MMMU-Pro and F1 score on POPE. We evaluate $500$ items per benchmark.}
\label{tab:general}
\begin{tabular}{lcccc}
\toprule
& MMStar & POPE & MMMU & MMMU-Pro\\\midrule
Base & 66.4 & 88.0 & 55.1 & 39.4 \\
SFT & 63.6 & 88.5 & 56.5 & 40.8 \\
LocPos & 64.4 & 87.2 & 54.4 & 41.0 \\
Ours (1:10) & 65.4 & 88.6 & 56.0 & 41.0 \\
Ours (2:10) & 65.0 & 89.3 & 54.1 & 38.6 \\
Ours (10:10) & 63.8 & 88.5 & 56.2 & 39.2 \\
\bottomrule
\end{tabular}
\end{table}

Finally, we examine whether \look{} training affects the model's general capabilities. As shown in~\cref{tab:general}, on MMStar~\citep{chen2024mmstar}, POPE~\citep{li2023pope}, MMMU~\citep{yue2024mmmu}, and MMMU-Pro~\citep{yue2025mmmupro}, all of our models remain within $2.5$ points of the SFT baseline. In particular, Ours (1:10), which uses the smallest amount of \look{} data while already substantially reducing hallucination in the absence of visual input, performs on par with or better than SFT across all four benchmarks. We hypothesize that the transfer of visual awareness learned by \look{} to the general VLM benchmarks needs more careful datasets and training design. Also note that the visual hallucination can arise from the semantic perception and understanding levels which need extra techniques to address in addition to the \look{}. We leave this to the future work.

\section{Related work}
\label{sec:related_work}
\paragraph{Visual hallucination.}
Existing benchmarks such as CHAIR~\citep{rohrbach2018chair}, POPE~\citep{li2023pope}, ROPE~\citep{chen2024rope}, and HallusionBench~\citep{guan2024hallusionbench} measure different forms of mismatch between visual evidence and generated claims, and methods address such errors through instruction data~\citep{liu2024lrv}, preference alignment~\citep{sun2024llavarlhf}, or decoding~\citep{leng2024vcd,huang2024opera}. Other studies expose limits of visual representations~\citep{tong2024eyeswideshut}, elementary perception~\citep{rahmanzadehgervi2024blind}, and spatial relations~\citep{kamath2023whatsup}. Recently, \citet{asadi2026mirage} show the illusion even in frontier models where models describe images they were never given. VisABench turns this observation into a quantitative measure, while \look{} targets the underlying capability: knowing what visual input is available before reasoning about what it contains.

\paragraph{Self-supervised methods to mitigate hallucination.}
Context prediction~\citep{doersch2015context}, jigsaw puzzles~\citep{noroozi2016jigsaw}, rotation prediction~\citep{gidaris2018rotation}, and masked reconstruction~\citep{he2022mae} derive supervision from transformations of the input image. Recent multimodal post-training methods, such as Visual Jigsaw~\citep{wu2026visualjigsaw} and SSL-R1~\citep{xie2026sslr1}, extend this paradigm to language-model outputs and verifiable rewards. \look{} shares their annotation-free construction, but differs in asking the model to localize its available visual input within the coordinate frame of the original image. As a result, the task is solved through positional information rather than semantic cues from the image content.

\section{Limitations}
\label{sec:limitations}
Our work has several limitations. First, our study focuses primarily on a single model family. The current implementation of \look{} requires a variable visual-token grid and access to absolute positional information, so architectures that encode geometry differently may require alternative task designs. Second, VisABench replaces visual features with a constant embedding, which is far from the distribution of natural images. As a result, awareness of missing visual input does not necessarily imply faithful grounding on challenging real images. Finally, visual hallucination may arise at the \emph{Look}, \emph{Perceive}, and \emph{Understand} levels, and developing methods that jointly address failures across all three levels remains an open problem.

\section{Conclusion}
We study visual hallucination from a more fundamental perspective by asking whether a VLM is aware of the visual input available to it before reasoning about what that input contains. To structure this question, we introduce a three-level hierarchy of visual grounding---\emph{Look}, \emph{Perceive}, and \emph{Understand}---and focus on the first level, which remains largely untested by existing benchmarks. We propose the self-supervised \look{} task, where a model must localize its available visual region using positional information while the excluded image content is never encoded. Our experiments show that pretrained VLMs struggle with this task, whereas fine-tuning on automatically generated \look{} data substantially improves visual awareness with little effect on general VLM performance. We further introduce VisABench to test whether a model recognizes when visual evidence is absent. On VisABench, standard instruction tuning increases the no-image false-Yes rate to $31.2\%$, while \look{} fine-tuning reduces it to $0.8\%$ without degrading real-image performance. These results suggest that visual hallucination can arise as early as the \emph{Look} level and that explicitly training visual-input awareness provides a simple, annotation-free way to improve visual grounding before semantic perception and reasoning begin.

\clearpage

\subsection*{Reproducibility statement}
\cref{sec:task} specifies the \look{} construction and the positional restoration, \cref{sec:visabench} the VisABench construction and scoring, and \cref{sec:setup} the training and evaluation protocol. \cref{app:training} gives the instruction mixture and training recipe, \cref{app:prompts} the prompts, and \cref{app:verification} the implementation checks. Accompanying scripts rebuild every table and figure from saved predictions without model inference and record hashes of their inputs.

\subsection*{Ethics statement}
The study uses existing research datasets and collects no new human-subject data. Images may depict identifiable people; any release must respect the source datasets' access and licensing terms. Improved awareness of visual input does not guarantee factual or safe outputs, and the models should not be described as having human-like awareness.

\subsection*{AI assistance disclosure}
AI tools assisted with manuscript editing, analysis-code preparation, and figure layout. 

\section*{Acknowledgements}
We acknowledge the computational resources provided by the Aalto Science-IT project (Triton cluster) and CSC -- IT Center for Science, Finland. We also thank the researchers and maintainers who make public datasets, pretrained vision-language models, and open-source training and evaluation tools available to the research community, enabling the experiments presented in this work.

\bibliography{iclr2027_conference}
\bibliographystyle{iclr2027_conference}
\clearpage
\appendix

\setcounter{table}{0}
\setcounter{figure}{0}
\renewcommand{\thetable}{A\arabic{table}}
\renewcommand{\thefigure}{A\arabic{figure}}

\crefalias{section}{appendix}

\section{Training data and implementation details}
\label{app:training}
\begin{table}[ht]
\centering\small
\caption{The instruction mixture replay datasets. \look{} examples reuse its image pool.}
\begin{tabular}{lrl}
\toprule
Dataset & Rows & Task\\\midrule
VQAv2~\citep{goyal2017vqav2} & 12,000 & Visual question answering\\
TextVQA~\citep{singh2019textvqa} & 6,000 & Text reading and question answering\\
Visual7W~\citep{zhu2016visual7w} & 6,000 & Grounded question answering\\
A-OKVQA~\citep{schwenk2022aokvqa} & 5,000 & Knowledge-based question answering\\
ChartQA~\citep{masry2022chartqa} & 5,000 & Chart reasoning\\
TextCaps~\citep{sidorov2020textcaps} & 4,000 & Captioning with text\\
ScienceQA~\citep{lu2022scienceqa} & 4,000 & Science question answering\\
GeomVerse~\citep{kazemi2023geomverse} & 4,000 & Geometric reasoning\\\midrule
Total & 46,000 &\\\bottomrule
\end{tabular}
\end{table}
We train in bfloat16 using scaled dot-product attention and apply LoRA to all linear layers with rank $16$ and $\alpha=32$, while keeping the vision encoder and projector frozen. We use a per-device batch size of $1$ with $8$ gradient-accumulation steps across four GPUs, giving an effective batch size of $32$. Training runs for two epochs with a learning rate of $10^{-4}$, a warmup ratio of $0.05$, a maximum sequence length of $4096$, and a maximum image area of $1{,}003{,}520$ pixels. We reserve $1\%$ of the combined training examples for validation.

\section{Implementation of the \look{} task}
\label{app:implementation}
\paragraph{Sampling the window.}
The processor resizes image $I$ to a canonical canvas of $R_w\times R_h$ pixels. For patch size $p$ and spatial merge factor $m$, each visual token corresponds to a merge block of side $c=pm$, so the canvas contains $B_w=R_w/c$ by $B_h=R_h/c$ blocks. We sample a rectangle $w=(b_{x_0},b_{y_0},b_{x_1},b_{y_1})$ on this block grid, with sides between $15\%$ and $80\%$ of the canvas, and serialize the target in the integer coordinate convention of Qwen3-VL:
\begin{equation}
 y(w)=1000\left(\frac{b_{x_0}}{B_w},\frac{b_{y_0}}{B_h},
                    \frac{b_{x_1}}{B_w},\frac{b_{y_1}}{B_h}\right).
\label{eq:target}
\end{equation}
Because each window is aligned with the visual-token grid, every token lies entirely inside or outside the selected region, making the target box exact up to coordinate rounding. We resize the full image before cropping and disable any subsequent resizing, so each crop patch is pixel-identical to the corresponding patch in the full image. The resulting crop features nevertheless differ from their full-image counterparts because the excluded patches do not participate in vision-encoder attention. We discard crops smaller than four merge blocks, which would otherwise be upsampled by the image loader. This removes $264$ of the $46{,}000$ generated training examples.

\paragraph{Restoring global positions.}
We override the three positional pathways through input arguments and forward hooks, without introducing new parameters or modifying the attention computation.
\begin{enumerate}[
    label=(\arabic*),
    labelindent=0pt,
    leftmargin=*,
    labelsep=0.5em,
    align=left
]
\item \textbf{Vision absolute positions.} Qwen3-VL interpolates a learned positional table to the current patch grid. We compute the interpolation indices and weights on the full-image grid and then select the entries corresponding to the crop patches.
\item \textbf{Vision rotary positions.} We offset the crop's two-dimensional patch indices by $(mb_{y_0},mb_{x_0})$. Because a shared translation cancels in the rotary query--key inner product~\citep{su2024rope}, this pathway does not by itself encode absolute location. We nevertheless restore it for consistency.
\item \textbf{Language-model positions.} We shift the height and width components of the image tokens' multimodal rotary positions by $(b_{y_0},b_{x_0})$, thereby preserving their spatial placement relative to the text tokens.
\end{enumerate}
In a controlled comparison, overriding the vision absolute positions changes the vision-encoder output by a relative $L_2$ distance of $0.676$, compared with only $0.033$ for the rotary-position override. On Qwen2.5-VL~\citep{bai2025qwen25vl}, whose vision encoder uses only rotary positional encoding, restoring the crop positions leaves the output unchanged. This further motivates our requirement for an absolute positional signal in the \look{} task.

\section{VisABench details}
\label{app:visabench}
We use the prompt ``Does this image contain: \{expression\}? Answer with Yes or No.'', inserting the raw RefCOCO expression. Images are resized to the model's canonical geometry, consistent with our other evaluations. In the no-image condition, the image is still loaded so that the number of visual tokens matches the real-image input, but the vision encoder is bypassed. The feature function instead returns the \texttt{<|image\_pad|>} embedding at every visual-token position together with all-zero DeepStack features. For the wrong-image condition, we sort items by image and pair each item with the expression from the item halfway around the list, skipping pairs that originate from the same image. The sets $\mathcal Y$ and $\mathcal N$ include all vocabulary tokens whose decoded surface forms, after removing surrounding whitespace and ignoring case, are ``yes'' and ``no'', respectively. The two sets are disjoint by construction. Each model is evaluated with one forward pass per item and condition, for a total of $32{,}256$ forward passes per model.

\section{Evaluation details}
\label{app:downstream}
\paragraph{Protocol.}
MMStar~\citep{chen2024mmstar}, POPE~\citep{li2023pope}, and MMMU-Pro~\citep{yue2025mmmupro} are capped at $500$ items; MMMU~\citep{yue2024mmmu} reports the $425$ validation items within its cap. The harness is VLMEvalKit~\citep{duan2024vlmevalkit}, and POPE reports F1.

\paragraph{Answer extraction.}
Without an LLM judge, the evaluation harness marks a multiple-choice response as incorrect whenever its answer-extraction rule fails, which disproportionately penalizes verbose outputs. Under the default extraction, the pretrained model obtains an MMMU score of only $0.089$, largely because its chain-of-thought responses are not successfully parsed. We therefore re-score all saved outputs using the same fixed sequence of extraction rules for every model: the last explicit answer phrase, a bolded option letter, the harness's option rule applied to the final non-empty line, and a unique match to an option string within the final $300$ characters. Responses that remain unresolved are counted as incorrect. For open-ended questions, we match the normalized final answer exactly or as a substring, using a $1\%$ relative tolerance for numerical answers. With this corrected scoring procedure, the pretrained model's MMMU score increases to $0.551$. This change reflects improved answer extraction rather than any modification to the model or its outputs.

\section{Prompt inventory}
\label{app:prompts}
The evaluation cycles through $11$ groups and samples one prompt within each group. Every prompt contains a format cue and asks for the visible region, not a description of its content. The inventory below is copied from the evaluation code.
\input{analysis/prompts.tex}

\section{Verifying the positional restoration}
\label{app:verification}
We validate positional restoration independently of downstream performance. Translating the crop patch indices by the grid offset must exactly reproduce the corresponding rows of the full-image position indices. The same row mapping is used to select the absolute interpolation entries, so any indexing error affects both pathways and is detected by this check. We additionally use a painted-patch test to verify the processor's merge-block-major flattening order. After canonical resizing, cropping must leave the retained patch inputs unchanged, and the loader must not upsample the crop. During training, crop offsets must also survive the full dataset-loading pipeline. We therefore log the first active positional-override calls to detect any silent loss of these offsets.

The output parser supports both plain-text bounding boxes and the model's special-token box format. For all model comparisons, we verify that images, prompts, and target windows are identical, recompute IoU directly from the predicted coordinates, and record SHA-256 hashes for every input file. All permutation tests use fixed random seeds.

\end{document}

%% file: math_commands.tex
\usepackage{amsmath,amsfonts,bm}

\def\eqref#1{equation~\ref{#1}}

\def\1{\bm{1}}

\DeclareMathAlphabet{\mathsfit}{\encodingdefault}{\sfdefault}{m}{sl}
\SetMathAlphabet{\mathsfit}{bold}{\encodingdefault}{\sfdefault}{bx}{n}

%% file: analysis/prompts.tex
{\renewcommand{\arraystretch}{1.3}%
\begin{longtable}{p{0.13\textwidth}p{0.80\textwidth}}
\toprule Group & \normalfont\normalsize Prompt\\\midrule
\endhead
see & \promptbox{Which region of the full image can you see? Answer with a bounding box.} \\
 & \promptbox{What part of the image is visible to you? Give a bounding box.} \\
 & \promptbox{Identify the area you are able to see; respond with a bounding box.} \\
 & \promptbox{Report the visible region as a bounding box.} \\
look & \promptbox{Where in the full image are you looking? Answer with a bounding box.} \\
 & \promptbox{Which region are you looking at? Reply with box coordinates.} \\
 & \promptbox{Point out the region you are looking at, as a bounding box.} \\
 & \promptbox{Give the bounding box of the area you are currently looking at.} \\
observe & \promptbox{Which part of the image are you observing? Answer with a bounding box.} \\
 & \promptbox{Mark the region you can observe; return a bounding box.} \\
 & \promptbox{What area is under your observation? Provide a bounding box.} \\
perceive & \promptbox{Which region of the scene do you perceive? Answer with a bounding box.} \\
 & \promptbox{Delimit the portion of the image you perceive, in bounding-box form.} \\
 & \promptbox{State, as a bounding box, the region you can perceive.} \\
view & \promptbox{What is your field of view within the full image? Answer with a bounding box.} \\
 & \promptbox{Which region falls inside your view? Express it as a bounding box.} \\
 & \promptbox{Give the bounding box of your viewport.} \\
 & \promptbox{Which rectangle of the original image are you viewing? Answer with its bounding box.} \\
examine & \promptbox{Which region are you able to examine? Answer with a bounding box.} \\
 & \promptbox{Give the bounding box of the area you can inspect.} \\
discern & \promptbox{Which part of the image can you discern? Answer with a bounding box.} \\
 & \promptbox{What region can you make out? Give its bounding box.} \\
 & \promptbox{Which area do you notice? Answer with a bounding box.} \\
 & \promptbox{Which portion of the image can you glimpse? Provide a bounding box.} \\
access & \promptbox{Which region of the image is accessible to you? Answer with a bounding box.} \\
 & \promptbox{What portion of the image is available to you? Give a bounding box.} \\
origin & \promptbox{This is part of a larger image. Where does it come from? Answer with a bounding box.} \\
 & \promptbox{Locate this view within the full image; return a bounding box.} \\
 & \promptbox{Where is this region situated in the original image? Answer with box coordinates.} \\
 & \promptbox{Return the bounding box of this view in the full image.} \\
occlusion & \promptbox{Everything outside one rectangle is hidden from you. Which rectangle is not? Answer with a bounding box.} \\
 & \promptbox{Only one region of the image reaches you. Which one? Give its bounding box.} \\
terse & \promptbox{Visible region (bounding box):} \\
 & \promptbox{Your viewport, as a bounding box.} \\
 & \promptbox{Looking at (bounding box):} \\
 & \promptbox{Seen region, as a box:} \\
\bottomrule
\end{longtable}}